\documentclass[twocolumn,a4paper]{article}

\usepackage{amsmath,amssymb,amsfonts}
\usepackage{algorithmic}
\usepackage{graphicx}
\usepackage{placeins}
\usepackage{adjustbox}
\usepackage[dvipsnames]{xcolor}
\usepackage[
  colorlinks=true,
  breaklinks=true,
  urlcolor=MidnightBlue,
  linkcolor=MidnightBlue,
  citecolor=MidnightBlue
]{hyperref}
\usepackage{breakurl}
\usepackage{fancyvrb}
\usepackage{authblk}
\usepackage{orcidlink}

\usepackage{booktabs}
\usepackage[authoryear]{natbib}
\let\cite\citep
\usepackage[flushleft]{threeparttable}
\usepackage{balance}
\usepackage{flushend}

\usepackage[T1]{fontenc}

\begin{document}

\title{Cleaning English Abstracts of Scientific Publications\thanks{For able research assistance, we thank Siva Pawan Venna.}}
\date{}
\author[1]{Michael E. Rose \orcidlink{0000-0002-4128-4236}}
\author[2]{Nils A. Herrmann \orcidlink{0009-0000-9429-2873}}
\author[1]{Sebastian Erhardt \orcidlink{0000-0002-2933-6451}}
\affil[1]{Max Planck Institute for Innovation and Competition, Munich, Germany}
\affil[2]{Technical University Munich, Munich, Germany}

\maketitle

\begin{abstract}
Scientific abstracts are often used as proxies for the content and thematic focus of research publications. However, a significant share of published abstracts contains extraneous information—such as publisher copyright statements, section headings, author notes, registrations, and bibliometric or bibliographic metadata—that can distort downstream analyses, particularly those involving document similarity or textual embeddings. We introduce an open-source, easy-to-integrate language model designed to clean English-language scientific abstracts by automatically identifying and removing such clutter.  We demonstrate that our model is both conservative and precise, alters similarity rankings of cleaned abstracts and improves information content of standard-length embeddings.
\end{abstract}

\section{Problem Statement}
Abstracts of scientific publications are a common way to judge the paper's content. This is because they are in the public domain. Unfortunately, an abstract can be polluted through various things, which adversely affect the quality of subsequent analysis. Here we document a language model that cleans English-language abstracts.

Clutter text, as we call in it in this document, or meta information may enter the abstract because of publishers, because of journals, and because of authors. Clutter imposed by publishers include copyright statements (ex. "© 2020 Springer" or "All rights reserved") and order information (ex. "Payment must accompany order."). Clutter imposed by journals may include section titles and abstract templates (ex. "Abstract.", "Data \& Samples:" or "Conclusion-"), and keywords and codes (ex. "JEL Codes: O15"). Meta information imposed by the authors include registration information (ex. "ClinicalTrials.gov: NCT012345678"), translations (ex. "This article is a translation of \ldots") and funding statements (ex. "Funding: Funding was provided by \ldots"). Sometimes there are also internal references to figures or tables (ex. "(Fig. 1)") or references (ex. "[1-4]"). Finally, some abstracts include references with bibliometric information above the usual author-year combination. This includes the journal, arXiv IDs, or even the title of the cited article.

There are many reasons why researchers would want to remove such clutter. For instance, they affect the similarity of two documents. Two papers from the same journal would naturally appear more similar to each other just because they used the same template. The same holds true for the abstracts of two journals by the same publisher, which include its copyright statement. Another important reason is to remove clutter so that more of the relevant content can be used in vectors of fixed length. This is an important reason given the advent of BERT-based language models \citep{devlin_bert_2019}.

Our model is easy to use. In Python, users can include a simple pre-processing step in their pipeline via the \texttt{transformers} library \citep{wolf_transformers_2020}.

\section{Training}
\subsection{Training data}
Our model is trained and evaluated on 9,000 English abstracts of scientific publications since 1970 from the Scopus database. We started from a small random sample using publications from 2018, and gradually expanded the training set with abstracts that contain one of the above categories of clutter.

\begin{table}[]
    \caption{Distribution of fields among the training and validation data}
    \label{tab:dist_fields}
    \centering
    \begin{tabular}{lrr}
\toprule
 & Count & Share \\
Field (ASJC-2) &  &  \\
\midrule
Medicine & 2171 & 24.1 \\
Physics & 1039 & 11.5 \\
Engineering & 954 & 10.6 \\
Chemistry & 946 & 10.5 \\
General Social Sciences & 881 & 9.8 \\
Biochemistry & 861 & 9.6 \\
Computer Sciences & 799 & 8.9 \\
Materials Sci. & 724 & 8.0 \\
Mathematics & 559 & 6.2 \\
Biology & 497 & 5.5 \\
Chem. Engineering & 469 & 5.2 \\
Arts/Humanities & 461 & 5.1 \\
Environ. Sci. & 425 & 4.7 \\
Earth and Planetary Sciences & 347 & 3.9 \\
Business & 283 & 3.1 \\
Energy & 281 & 3.1 \\
Psychology & 220 & 2.4 \\
Economics & 219 & 2.4 \\
Pharmaceutics & 204 & 2.3 \\
Immunology & 201 & 2.2 \\
Neuroscience & 194 & 2.2 \\
Decision Sciences & 122 & 1.4 \\
Health & 109 & 1.2 \\
Nursing & 104 & 1.2 \\
Multidisciplinary & 103 & 1.1 \\
Veterinary & 49 & 0.5 \\
Dentistry & 32 & 0.4 \\
\bottomrule
\end{tabular}

\end{table}

Table \ref{tab:dist_fields} presents the distribution over the various academic fields. Fields correspond to Elsevier's All-Science Journal Classification (ASJC) at the top level, which are devoted to journals and conference proceeding. A journal or conference proceeding has up to 7 fields, but most have just one, as \cite{hottenrott_rise_2021} show. The distribution of fields of our dataset corresponds to the overall distribution of all papers published in 2024, according to Scopus.

Table \ref{tab:dist_years} presents the distribution of publication years within the dataset. The over-representation of 2018 is due to the sampling strategy.

\begin{table}[]
    \caption{Distribution of publication years among the training and validation data}
    \label{tab:dist_years}
    \centering
    \small
    \begin{tabular}{lrr}
\toprule
 & Count & Share \\
Year &  &  \\
\midrule
1970 & 17 & 0.2 \\
1971 & 14 & 0.2 \\
1972 & 15 & 0.2 \\
1973 & 34 & 0.4 \\
1974 & 38 & 0.4 \\
1975 & 30 & 0.3 \\
1976 & 38 & 0.4 \\
1977 & 44 & 0.5 \\
1978 & 38 & 0.4 \\
1979 & 48 & 0.5 \\
1980 & 39 & 0.4 \\
1981 & 36 & 0.4 \\
1982 & 43 & 0.5 \\
1983 & 59 & 0.7 \\
1984 & 56 & 0.6 \\
1985 & 61 & 0.7 \\
1986 & 58 & 0.6 \\
1987 & 73 & 0.8 \\
1988 & 68 & 0.8 \\
1989 & 60 & 0.7 \\
1990 & 68 & 0.8 \\
1991 & 67 & 0.7 \\
1992 & 81 & 0.9 \\
1993 & 77 & 0.9 \\
1994 & 89 & 1.0 \\
1995 & 78 & 0.9 \\
1996 & 91 & 1.0 \\
1997 & 106 & 1.2 \\
1998 & 123 & 1.4 \\
1999 & 117 & 1.3 \\
2000 & 150 & 1.7 \\
2001 & 143 & 1.6 \\
2002 & 121 & 1.3 \\
2003 & 150 & 1.7 \\
2004 & 142 & 1.6 \\
2005 & 173 & 1.9 \\
2006 & 213 & 2.4 \\
2007 & 208 & 2.3 \\
2008 & 219 & 2.4 \\
2009 & 233 & 2.6 \\
2010 & 236 & 2.6 \\
2011 & 257 & 2.9 \\
2012 & 287 & 3.2 \\
2013 & 270 & 3.0 \\
2014 & 252 & 2.8 \\
2015 & 509 & 5.7 \\
2016 & 316 & 3.5 \\
2017 & 316 & 3.5 \\
2018 & 363 & 4.0 \\
2019 & 1188 & 13.2 \\
2020 & 361 & 4.0 \\
2021 & 273 & 3.0 \\
2022 & 283 & 3.1 \\
2023 & 279 & 3.1 \\
2024 & 293 & 3.3 \\
\bottomrule
\end{tabular}

\end{table}

We manually labeled all the abstracts through an interactive user interface. That is, we marked the text that is supposed to be dropped. Marked text belongs to the category REM (for remove). The test-training split is 20\% to 80\%.


\subsection{Training setup}

The model was trained with spacy 4.0 \cite{honnibal_spacy_2023}. spacy conducted Named Entity Recognition for the labeled text and is thus able to predict membership to this category.

We ran the trainer for up to 100 epochs with an early stopping when the F1 score doesn't improve more than 0.0001 upon the best model for 20 epochs. The model finished after 52 epochs. On four NVIDIA A100-SXM4-40GB315 GPUs, the training took 4 hours with an effective batch size of 16.

\begin{figure}
    \caption{Loss and performance metrics per epoch}
    \label{fig:line-training}
    \centering
    \includegraphics[width=\linewidth]{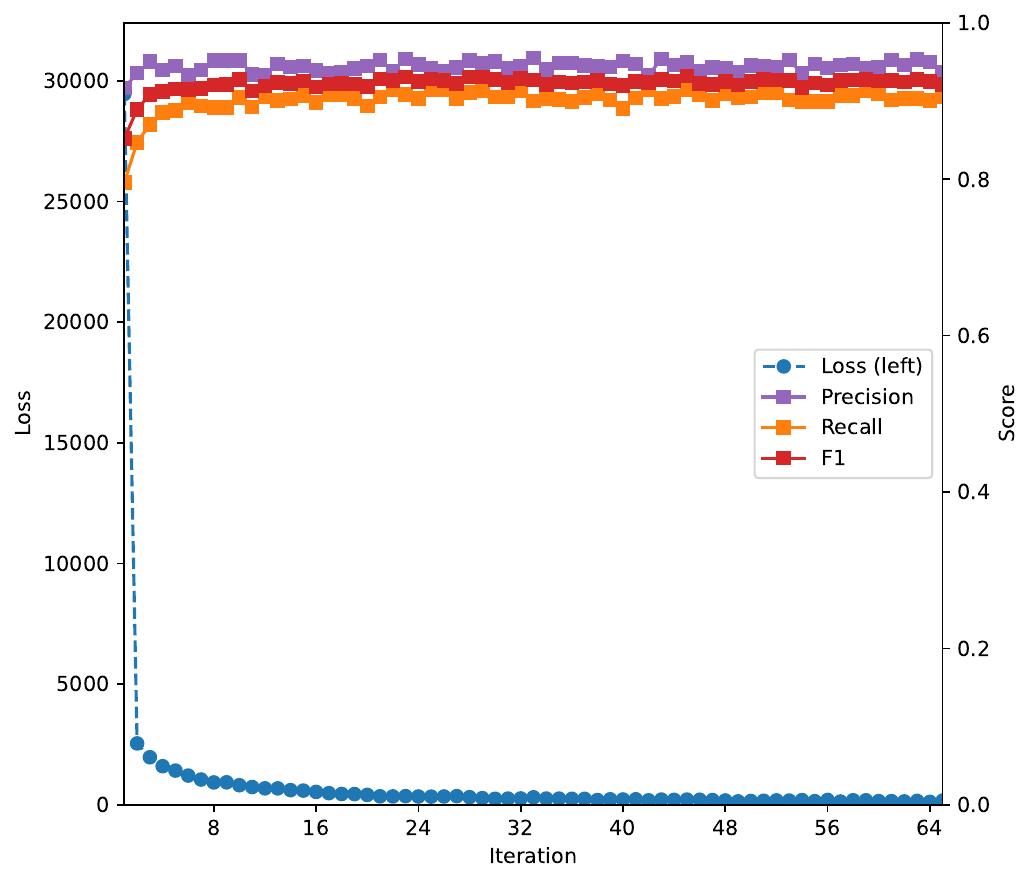}
\end{figure}

Figure \ref{fig:line-training} shows the evolution of the loss function (left scale) and performance metrics (right scale). While loss quickly plummeted, the training evaluation metrics improve steadily but marginally.

We perform token-level evaluation on a hold-out set consisting of 20\% of the data. The resulting precision is 0.973, the recall is 0.919 and the F1-score equals 0.945.

\FloatBarrier

\subsection{Evaluation}

Our evaluation is hampered by the lack of an external ground truth. Therefore we rely on two samples to show how the Abstract Cleaner works.

\begin{table*}[]
    \caption{Evaluation of Abstract Cleaner within training sample}
    \label{tab:eval_training}
    \centering
    \begin{tabular}{lrrrrrr}
\toprule
 & \multicolumn{2}{c}{} & \multicolumn{2}{c}{Excess} & \multicolumn{2}{c}{Missing} \\
 & Count & Share correct & Share pred. & \# tokens & Share pred. & \# tokens \\
Has labels &  &  &  &  &  &  \\
\midrule
No & 4044 & 99.93\% & 0.07\% & 6.33 &  &  \\
Yes & 4956 & 96.97\% & 0.56\% & 7.96 & 2.26\% & 7.31 \\
\bottomrule
\end{tabular}

\end{table*}

In Table \ref{tab:eval_training} we present the effect on the training set itself. We show how many abstracts have been affected, and whether normal text was erroneously removed (``Excess'') or whether clutter was mistakenly not removed (``Missing''). We do so separately for abstracts that have clutter and those that have not.

In general, the Abstract Cleaner works very conservatively. In less than 0.1\% of cases, where the abstract has no labeled clutter, the Abstract Cleaner removes text. On average, it removes 6.33 tokens for these abstracts. The same holds true for abstracts with labeled clutter, where we find an excess removal of 7.96 tokens on average. Manual inspection reveals that these excess tokens are usually punctuation and numbers.

At the same time, it achieves a high precision, as most abstracts with labeled clutter are correctly affected. When it removes clutter, it tends to remove too little, too: in 2.26\% of cases it misses tokens, namely 7.31 on average.

\begin{figure}
    \caption{Relation abstract length and misses/excess}
    \label{fig:bar_predictions}
    \centering
    \includegraphics[width=\linewidth]{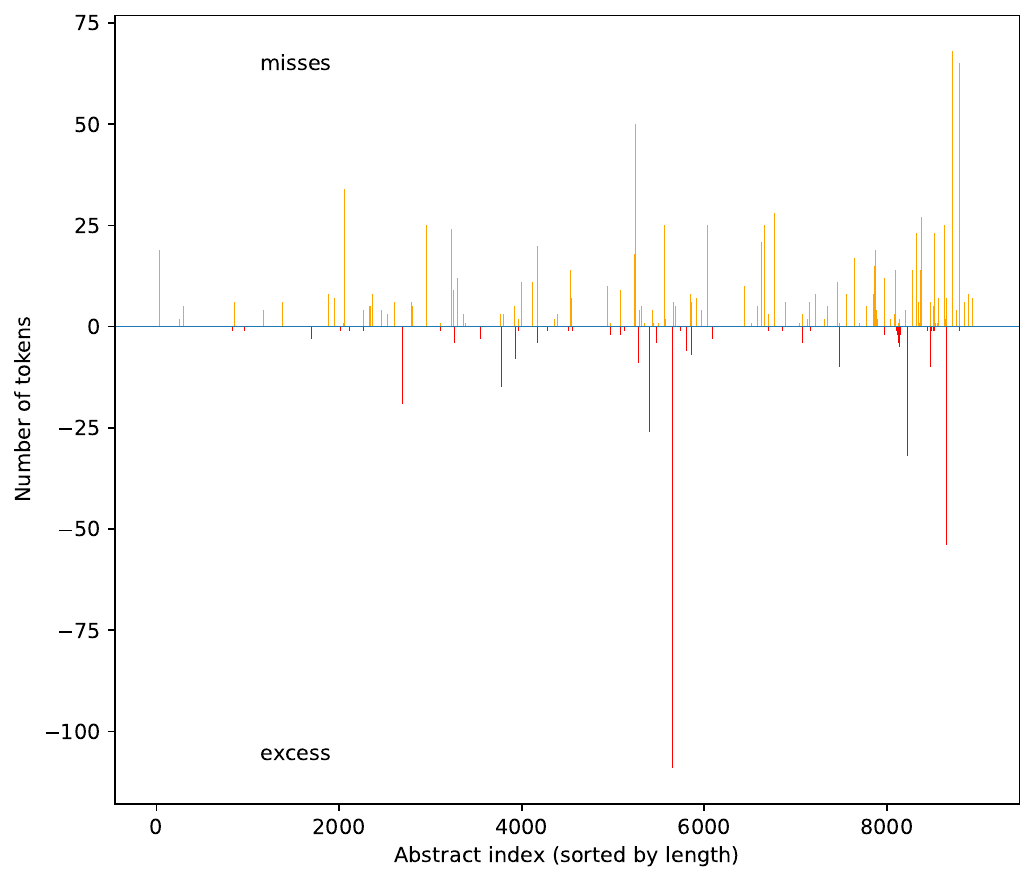}
    
\end{figure}

Abstracts have different length however. Working with scientific abstracts it is more worrisome if text from short abstracts is erroneously removed than from long abstracts. Figure \ref{fig:bar_predictions} shows that for the training set, the propensity to observe excess removal of tokens increases with abstract length (lower part). The number of missed tokens also grows with abstract length (upper part).

To see how the Abstract Cleaner behaves on unseen data, we generate four training sets of 100 abstracts each. These were selected by regular expression from Scopus. Each training set is meant to represent one form of clutter, ideally in a pure form: Long or full-text citations, copyright statements, other clutter (keywords, section titles, study registration information, etc.). We complement this with a selection of random abstracts that have neither of these clutter.

\begin{table*}
    \caption{Evaluation of Abstract Cleaner within training sample}
    \label{tab:eval_test}
    \centering
    \begin{tabular}{lrrrrrr}
\toprule
 & \multicolumn{2}{c}{} & \multicolumn{2}{c}{Excess} & \multicolumn{2}{c}{Missing} \\
 & Count & Share correct & Share pred. & \# tokens & Share pred. & \# tokens \\
\midrule
Citations & 100.00 & 55.00\% & 20.00\% & 2.50 & 33.00\% & 8.18 \\
Copyright & 100.00 & 96.00\% & 1.00\% & 11.00 & 3.00\% & 15.67 \\
Other & 100.00 & 84.00\% & 2.00\% & 9.50 & 14.00\% & 15.43 \\
Random & 100.00 & 98.00\% & 0.00\% &  & 2.00\% & 5.00 \\
\bottomrule
\end{tabular}

\end{table*}

Table \ref{tab:eval_test} presents the results of these four exercises, again split by excess prediction and missing predictions. For abstracts with citations, only 55\% of abstracts are correctly cleaned. In 20\% of cases, the Abstract Cleaner removes too much, namely 2.5 tokens on average. It misses to remove tokens in 33\% of cases, namely 8.18 tokens on average. In the copyright sample, the Abstract Cleaner is precisely correct 96\& of the time. Only 1\% of abstracts looses information (11 tokens on average), while only 3\% of abstracts retain clutter (15.67 tokens on average). Concerning other clutter categories, 84\% of abstracts are correctly cleaned. 2\% of abstracts show excess removal of tokens (9.5 on average), and in 14\% of abstracts the Abstract Cleaner misses to remove clutter. Finally, among the 100 random abstracts, there is no single excess prediction. Only 2\% of abstracts retain clutter, namely 5 tokens on average.

\FloatBarrier

\section{Validation}

Having shown that the Abstract Cleaner reliably and conservatively removes clutter from scientific abstracts, we now show that cleaned abstracts affect relevant outcomes. 





To show the effect of the Abstract Cleaner, we compare the similarity ranking of articles before and after abstract cleaning. That is, we compute SPECTER2 embeddings representing original and cleaned abstracts, and compare the resulting similarity ranking relative to a focal patent.

To make the analysis comparable, we study a paper's references. For our testing papers, we retrieve their abstract as well as the abstracts of cited papers. Then we compare how often the ranking according to the cosines of the SPECTER2 embeddings with the cited papers changes due to cleaning.

For testing purpose, we select some of the recent Nobel awarding papers quoted in the Nobel committee's press statements. 
Specifically, we study 17 papers that awarded its authors a Nobel prize in either Medicine, Physics or Chemistry in 2025. 

We find that in all cases, the cosine similarity ranking changes. In 4 out of the 17 cases, the most similar reference differs when we clean the abstracts.




\FloatBarrier

\section{Usage}
The model and training data are publicly available on Huggingface:
\begin{itemize}
    \item \textbf{Model:}\\ \url{https://huggingface.co/mpi-inno-comp/abstract-cleaner}
    \item \textbf{Training data set:} \\
    \url{https://huggingface.co/datasets/mpi-inno-comp/abstract_dataset}
\end{itemize}

The cleaning of an abstract takes about 0.5 seconds on a 2.1 GHz GPU.

The model predicts which text belongs to the category REM (for remove) and which does not. It returns the start and the end indices. Equipped with these, the user has to slice the original text.

In python, we recommend the function shown in source code \ref{code:basequery}.

\paragraph{Example code for decluttering a short text\label{code:basequery}}
\begingroup
    \fontsize{5pt}{8pt}\selectfont
    \begin{Verbatim}[frame=single] 
import spacy
from spacy.util import filter_spans

cleaner = spacy.load("mpi-inno-comp/abstract-cleaner")

def clean_abstract(text: str | None, model) -> str | None:
    """Declutter scientific abstract."""
    original = text
    doc = model(text)
    if not doc.ents:
        return text.strip()
    # Build cleaned text by character slicing
    parts = []
    last_pos = 0
    for ent in doc.ents:
        parts.append(text[last_pos:ent.start_char])
        last_pos = ent.end_char
    parts.append(text[last_pos:])
    cleaned = "".join(parts).strip()    
    return cleaned if cleaned else text


original = "Lorem ipsum dolor sit amnet"
cleaned = clean_abstract(original, cleaner)
            
    \end{Verbatim}  
\endgroup

\addcontentsline{toc}{section}{References}
\bibliographystyle{apalike}
\bibliography{references.bib}

\end{document}